\documentclass[lettersize,journal]{IEEEtran}
\usepackage{amsmath,amsfonts}
\usepackage{amsthm}
\usepackage{amssymb}
\usepackage{algorithmic}
\usepackage{array}
\usepackage{booktabs}
\usepackage{textcomp}
\usepackage{stfloats}
\usepackage{url}
\usepackage{verbatim}
\usepackage{graphicx}
\usepackage{pifont}
\usepackage{threeparttable}
\usepackage{tabularx}
\usepackage{soul}
\usepackage{xcolor}
\usepackage{longtable}
\usepackage{mathtools}
\usepackage{orcidlink}
\usepackage{hyperref}
\usepackage{cleveref} 
\usepackage{subcaption}  
\usepackage{graphicx}
\usepackage{enumitem}

\usepackage{fontawesome}
\usepackage{academicons}

\newtheorem{remark}{Remark}

\def\BibTeX{{\rm B\kern-.05em{\sc i\kern-.025em b}\kern-.08em
    T\kern-.1667em\lower.7ex\hbox{E}\kern-.125emX}}
\usepackage{balance}
\begin{document}
\title{Constraint-Free Static Modeling of Continuum Parallel Robot}
\author{Lingxiao Xun, Matyas Diezinger, 	
Azad Artinian, Guillaume Laurent, Brahim Tamadazte
\thanks{   }}


\maketitle

\begin{abstract}
Continuum parallel robots (CPR) combine rigid actuation mechanisms with multiple elastic rods in a closed-loop topology, making forward statics challenging when rigid--continuum junctions are enforced by explicit kinematic constraints. Such constraint-based formulations typically introduce additional algebraic variables and complicate both numerical solution and downstream control. This paper presents a geometric exact, configuration-based and constraint-free static model of CPR that remains valid under geometrically nonlinear, large-deformation and large-rotation conditions. Connectivity constraints are eliminated by kinematic embedding, yielding a reduced unconstrained problem. Each rod of CPR is discretized by nodal poses on \(SE(3)\), while the element-wise strain field is reconstructed through a linear strain parameterization. A fourth-order Magnus approximation yields an explicit and geometrically consistent mapping between element end poses and the strain. Rigid attachments at the motor-driven base and the end-effector platforms are handled through kinematic embeddings. Based on total potential energy and virtual work, we derive assembly-ready residuals and explicit Newton tangents, and solve the resulting nonlinear equilibrium equations using a Riemannian Newton iteration on the product manifold. Experiments on a three-servomotor, six-rod prototype validate the model by showing good agreement between simulation and measurements for both unloaded motions and externally loaded cases.
\end{abstract}

%
\section{INTRODUCTION}
\label{sec:intro}
Slender elastic rods are a core primitive in continuum and soft robotics, enabling large bending and torsion for dexterous, safe interaction. Many design and control tasks require accurate, efficient {forward statics}: given actuation and external loads, predict the equilibrium shape and end-effector pose. Rod models strike a practical balance between low-fidelity kinematics and high-dimensional 3D continuum simulation, and are widely used for continuum-robot control and estimation~\cite{armanini2023soft}.

Cosserat rod theory provides a geometrically exact model of slender structures by representing the configuration as a curve on \(SE(3)\) and the strain as a body twist in \(\mathfrak{se}(3)\). Preserving this Lie-group structure is essential: under finite rotations, naive additive interpolation and updates can break objectivity, causing spurious stiffness and poor numerical behavior. These issues are well documented in geometrically exact beam/rod formulations and their finite-element implementations \cite{simo1985finite, simo1991geometrically, jelenic1999geometrically}, motivating Lie-group-consistent discretizations that couple translations and rotations for robust large-deformation simulation~\cite{sonneville2014geometric}.

Two complementary discretization paradigms dominate the continuum-robot literature. {Configuration-based} methods
take nodal poses as primary unknowns, as in Lie-group finite elements and isogeometric analysis
\cite{weeger2017isogeometric}. {Discrete elastic rods (DER)} discretize the centerline and elastic
energy via discrete frames \cite{bergou2008discrete}. Widely adopted in computer graphics, DER-type formulations have
more recently been explored in robotics for large deformations and rotations \cite{goldberg2019planar}. These
approaches inherit strong geometric consistency and can capture rich deformation patterns, but may require many
degrees of freedom and careful treatments to mitigate locking in slender regimes
\cite{sonneville2014geometric, weeger2017isogeometric}.

By contrast, {strain-based} formulations treat strain parameters as generalized coordinates,
yielding compact models that are well suited to real-time simulation and control. Representative examples include
piecewise-constant and piecewise-linear strain approximations, as well as geometric variable-strain models
\cite{8500341, li2023piecewise, 9057619}. While highly effective for single-backbone and open-chain structures, their
base-to-tip strain integration and boundary-value character make extensions to interconnected rods, branching, or
closed loops substantially more involved, often requiring additional constraint coupling or iterative projection
\cite{armanini2021discrete}.

Moving from an isolated rod to an assembly fundamentally changes the modeling problem: junction kinematics and
force/moment balance couple multiple elastic elements through rigid bodies and loop-closure conditions. Continuum
parallel robots are a representative and practically relevant example: multiple compliant rods connect a
rigid base to a rigid end-effector platform, forming a closed-loop mechanism that combines compliance with the load
capacity and accuracy advantages of parallel architectures~\cite{huang2022kinematic}. Existing CPR modeling work commonly couples multiple
Cosserat rods through explicit platform constraints and solves the resulting coupled boundary-value problem for
kinetostatics and force sensing~\cite{black2017parallel, orekhov2017modeling, rong2026kinetostatic}. Surveys further highlight that constraint handling and solver structure remain central bottlenecks as designs grow in complexity and as control-oriented computation is required online~\cite{lilge2024parallel}. In particular, enforcing rigid--continuum connectivity via explicit constraints
often increases the system size~\cite{armanini2021discrete, lilge2022kinetostatic} (e.g., through Lagrange multipliers), complicates numerical conditioning, and obscures a modular element-wise assembly structure.

This paper develops a compact, Lie-group-consistent forward-statics model for continuum parallel robots without explicit connection constraints. We use rod nodal poses in \(SE(3)\) as primary variables and locally reconstruct an element-wise strain field, thereby enforcing closed-loop connectivity via rigid-body kinematics. The resulting formulation provides an assembly-ready equilibrium residual and an explicit Newton tangent for efficient iterations on the product manifold of pose and strain variables. Experiments on a three-servomotor, six-rod prototype validate the model under both unloaded and externally loaded conditions.

The remainder of the paper is organized as follows. Section~\ref{sec.ps} states the problem and modeling assumptions. Section~\ref{sec.element} presents the configuration-based discretization and strain reconstruction. Section~\ref{sec:kin_variations} derives the kinematic variations and differential mappings for rigid embeddings. Section~\ref{sec:statics_opt} formulates forward statics as manifold optimization and presents the solver. Section~\ref{sec:experiments} reports experimental validation results, and Section~\ref{sec.conc} concludes the paper.
%
\section{PROBLEM STATEMENT} \label{sec.ps}
\begin{figure}[h]
	\centering
    \includegraphics[width=0.45\textwidth]{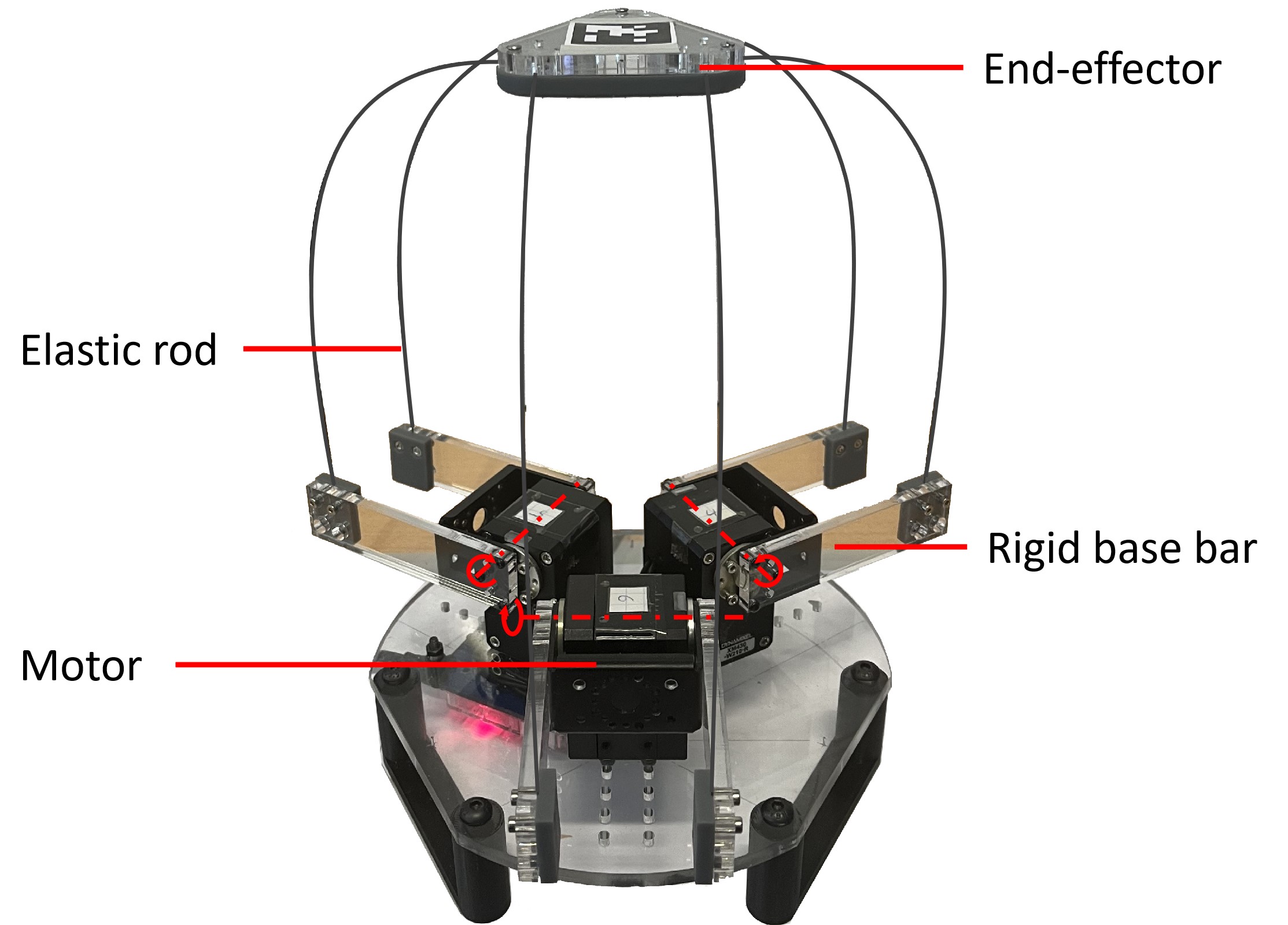}
	\caption{Continuum parallel robot considered in this work.}
	\label{fig:parallelrobot}
\end{figure}

We study the forward statics of a continuum parallel robot that combines rigid components with multiple elastic rods arranged in a closed-loop topology (Fig.~\ref{fig:parallelrobot}). The CPR in this work comprises a rigid base bar, a rigid end-effector, and multiple slender elastic rods connected through spatial attachment frames. Common modeling strategy treats rigid bodies and rods as separate subsystems and enforces their connections via explicit kinematic constraints. In closed-loop structures, this typically introduces additional algebraic equations (e.g., Lagrange multipliers), increases system size, and complicates both numerical solution and control.

Our goal is to develop a compact, control-friendly static formulation where rigid--continuum connectivity is enforced by construction. Each rod is discretized using nodal poses on \(SE(3)\), yielding a uniform kinematic representation that naturally accommodates closed-loop topologies without introducing additional connection constraints.
\section{CONFIGURATION-based DISCRETIZATION} \label{sec.element}
%
In this section, we introduce the discrete formulation for each CPR limb. Each limb is modeled as a slender elastic rod, modeled as a Cosserat rod with configuration \(g(s)\in SE(3)\), \(s\in[0,L]\), where
$$g(s)=\begin{bmatrix}R(s)&p(s)\\0&1\end{bmatrix}.$$ 
$R(s)\in SO(3)$ denotes the rotation matrix and $p(s)\in \mathbb{R}^3$ denotes the position vector. $s$ is the arc length of the limb and $L$ is the total length of the limb.

The body strain is defined by the left-trivialized derivative w.r.t. $s$
\begin{equation}
\widehat{\xi}(s)=g(s)^{-1}g'(s)\in\mathfrak{se}(3)\simeq\mathbb{R}^6,
\label{eq:cosserat_strain}
\end{equation}
$\widehat{(\cdot)}$ denotes the skew-symmetric transformation from $\mathbb{R}^6$ to $\mathfrak{se}(3)$.

\begin{figure}[h]
	\centering
    \includegraphics[width=0.35\textwidth]{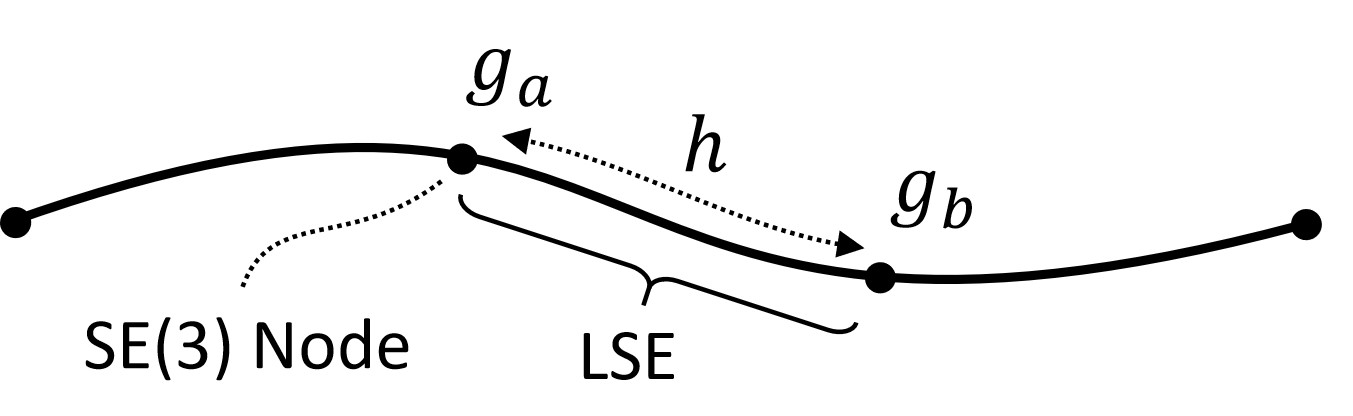}
	\caption{An elastic rod discretized into \(SE(3)\) nodes with the linear strain element (LSE).}
	\label{fig:rod}
\end{figure}

We discretize each rod using nodal poses on \(SE(3)\) and reconstruct an element-wise strain field through a linear strain parameterization, as illustrated in Fig.~\ref{fig:rod}. This configuration-based discretization leads to simple junction kinematics and naturally supports closed-loop topologies. In the following, we detail the linear strain element formulation and the resulting explicit mapping between element end poses and strains based on a Magnus expansion.
%
\subsection{Linear Strain Element (LSE)} \label{subsec:lse}
As shown in Fig.~\ref{fig:rod}, consider a single element of arc length \(h\), parameterized by the coordinate \(s\in[0,h]\). The spatial strain field within this element is assumed to vary linearly as
\begin{equation}\label{eq:linear_strain}
    \xi(s) = \bar{\xi} + (s - \tfrac{h}{2})\,\beta,
\end{equation}
where \(\bar{\xi}\in\mathbb{R}^6\) represents the mean strain, and \(\beta\in\mathbb{R}^6\) denotes the strain slope that captures the local variation of deformation along the element. This relation provides an element-wise parameterization of the strain distribution and will be used to construct a consistent mapping between nodal poses and strain parameters.
%
\subsection{Fourth-order Magnus Approximation} \label{subsec:magnus4}
Integrating the kinematics \(g'(s)=g(s)\widehat{\xi}(s)\) over one element \(s\in[0,h]\) relates the end poses
\(g_a=g(0)\) and \(g_b=g(h)\) through a group exponential:
\begin{equation}
g_a^{-1}g_b=\exp\!\big(\widehat{\Omega}(h)\big),\qquad \Omega(h)\in\mathbb{R}^6,
\label{eq:rel_pose_exp}
\end{equation}
where \(\widehat{(\cdot)}:\mathbb{R}^6\rightarrow\mathfrak{se}(3)\) is the hat operator. The integrated twist
\(\widehat{\Omega}(h)\) can be expressed by the Magnus expansion~\cite{blanes2009magnus}.

To obtain a closed-form approximation that is accurate up to \(\mathcal{O}(h^5)\), we use the fourth-order truncation in a compact moment form~\cite{blanes2006fourth}:
\begin{equation}
\widehat{\Omega}(h)\;\approx\;M_0 - [M_1,M_0],
\label{eq:magnus4_moment_compact}
\end{equation}
with the two moments defined as
\begin{equation}
M_0=\int_{0}^{h}\widehat{\xi}(s)\,ds,
\qquad
M_1=\frac{1}{h}\int_{0}^{h}\Big(s-\frac{h}{2}\Big)\widehat{\xi}(s)\,ds,
\label{eq:magnus_moments}
\end{equation}
and \([\cdot,\cdot]\) the Lie bracket in \(\mathfrak{se}(3)\).

Under the linear strain assumption \(\xi(s)=\bar\xi+\big(s-\tfrac{h}{2}\big)\beta\), the moments in~\eqref{eq:magnus_moments} can be evaluated directly.

Using \(\int_{0}^{h}(s-\tfrac{h}{2})\,ds=0\) and \(\int_{0}^{h}(s-\tfrac{h}{2})^2 ds=\tfrac{h^3}{12}\), we obtain
\begin{equation}
M_0=h\,\widehat{\bar\xi},
\qquad
M_1=\frac{h^2}{12}\,\widehat{\beta}.
\label{eq:magnus_moments_closed}
\end{equation}

Substituting \eqref{eq:magnus_moments_closed} into \eqref{eq:magnus4_moment_compact} yields
\begin{equation}
\widehat{\Omega}(h)\;\approx\;h\,\widehat{\bar\xi}-\frac{h^3}{12}\,[\widehat{\beta},\widehat{\bar\xi}].
\label{eq:magnus4_hat}
\end{equation}

Using the vector Lie bracket \([\beta,\bar\xi]\in\mathbb{R}^6\) induced by \(\mathfrak{se}(3)\),
equivalently \([\beta,\bar\xi]=\mathrm{ad}_{\beta}\bar\xi\), and denoting the identity matrix as $\mathbb{I}$, we arrive at the compact vector form
\begin{equation}\label{eq:magnus4}
\Omega(h)
=\Big(h\,\mathbb{I}-\frac{h^3}{12}\mathrm{ad}_{\beta}\Big)\bar\xi,
\end{equation}
which explicitly relates the integrated twist to the mean and slope of the strain distribution.

Finally, given the nodal poses \(g_a\) and \(g_b\) and the local slope \(\beta\), the mean strain \(\bar\xi\) can be retrieved without iteration by inverting \eqref{eq:magnus4}:
\begin{equation}\label{eq:explicit_meanstrain}
\bar{\xi} = A^{-1}\,\mathrm{Log}\!\left(g_a^{-1}g_b\right),
\qquad
A = h\mathbb{I} - \tfrac{h^{3}}{12}\mathrm{ad}_{\beta},
\end{equation}
where \(\mathrm{Log}(\cdot)\) is the logarithm map on \(SE(3)\).

This provides an explicit and geometrically consistent mapping between the nodal poses and the strain parameters \((\bar{\xi},\beta)\); the element geometry is fully determined by \(\{g_a,g_b,\beta\}\).

The closed-form relation derived in this section is the key enabler of an efficient variational formulation: it allows us to express the element strain variables directly in terms of nodal poses and \(\beta\), thereby enabling the computation of energy variations and equilibrium residuals without inner iterations. In the next section, we differentiate~\eqref{eq:explicit_meanstrain} under right perturbations and derive the kinematic Jacobians that map nodal and generalized-coordinate increments to the mean-strain variation \(\delta\bar\xi\).
%
\section{KINEMATIC VARIATIONS} \label{sec:kin_variations}
This section derives the differential kinematics needed for static equilibrium and tangent construction. Starting from
the closed-form mean-strain map~\eqref{eq:explicit_meanstrain}, we first obtain an element-level linearization that
relates \(\delta\bar\xi\) to perturbations of the two end poses and the slope parameter. We then incorporate the
rigid--continuum embeddings at the rod boundary nodes and introduce a sparse differential mapping that projects the
generalized increments \(\delta q\) to the local element increments. These relations provide the Jacobians used in the assembly of element residuals and Newton tangents in Section~\ref{sec:statics_opt}.
%
\subsection{Element-level Linearization}
\label{subsec:var_element}
For one LSE element with end poses \(g_a\) and \(g_b\), we use right perturbations
\(\delta\zeta_a,\delta\zeta_b\in\mathbb{R}^6\simeq\mathfrak{se}(3)\).

The variation of~\eqref{eq:explicit_meanstrain} yields
\begin{equation}\label{eq:linearization}
\delta\bar{\xi}
= J_1\,\delta\zeta_a
+ J_2\,\delta\zeta_b
+ J_3\,\delta\beta,
\end{equation}
where $J_{1}$, $J_{2}$ and $J_{3}$ are deduced as
\begin{align}
J_{1} &= A^{-1}\operatorname{dexp}^{-1}_{\Omega}\operatorname{Ad}_{\exp(-\Omega)}, \label{eq:J1}\\
J_{2} &= A^{-1}\operatorname{dexp}^{-1}_{\Omega},\label{eq:J2}\\
J_{3} &= \tfrac{1}{12}A^{-1}h^{3}\operatorname{ad}_{\bar{\xi}}. \label{eq:J3}
\end{align}
here \(\Omega=\mathrm{Log}(g_a^{-1}g_b)\).
Definitions of \(\operatorname{dexp}^{-1}\), \(\operatorname{Ad}\), and \(\operatorname{ad}\) are given in Appendix.

For compactness, we rewrite~\eqref{eq:linearization} as
\begin{equation}\label{eq:bx}
    \delta\bar{\xi}=B\,\delta x,
\end{equation}
with
\begin{equation}\label{eq:Be_def}
B := \begin{bmatrix} J_1 & J_2 & J_3 \end{bmatrix},
\quad
\delta x := \begin{bmatrix} \delta\zeta_a^\top & \delta\zeta_b^\top & \delta\beta^\top \end{bmatrix}^\top.
\end{equation}
%
\subsection{Boundary-node Embeddings} \label{subsec:boundary_embed}
Each rod \(k\in\{1,\dots,6\}\) is discretized by \(n_k\) LSE elements with nodal poses
\(\{g_{k,0},g_{k,1},\dots,g_{k,n_k}\}\subset SE(3)\), where element \(e\) uses
\begin{equation}
g_a^{k,e}=g_{k,e-1},\qquad g_b^{k,e}=g_{k,e}.
\label{eq:element_end_nodes}
\end{equation}
interior nodes \(g_{k,1},\dots,g_{k,n_k-1}\) are independent variables.

\begin{figure}[h]
	\centering
    \includegraphics[width=0.45\textwidth]{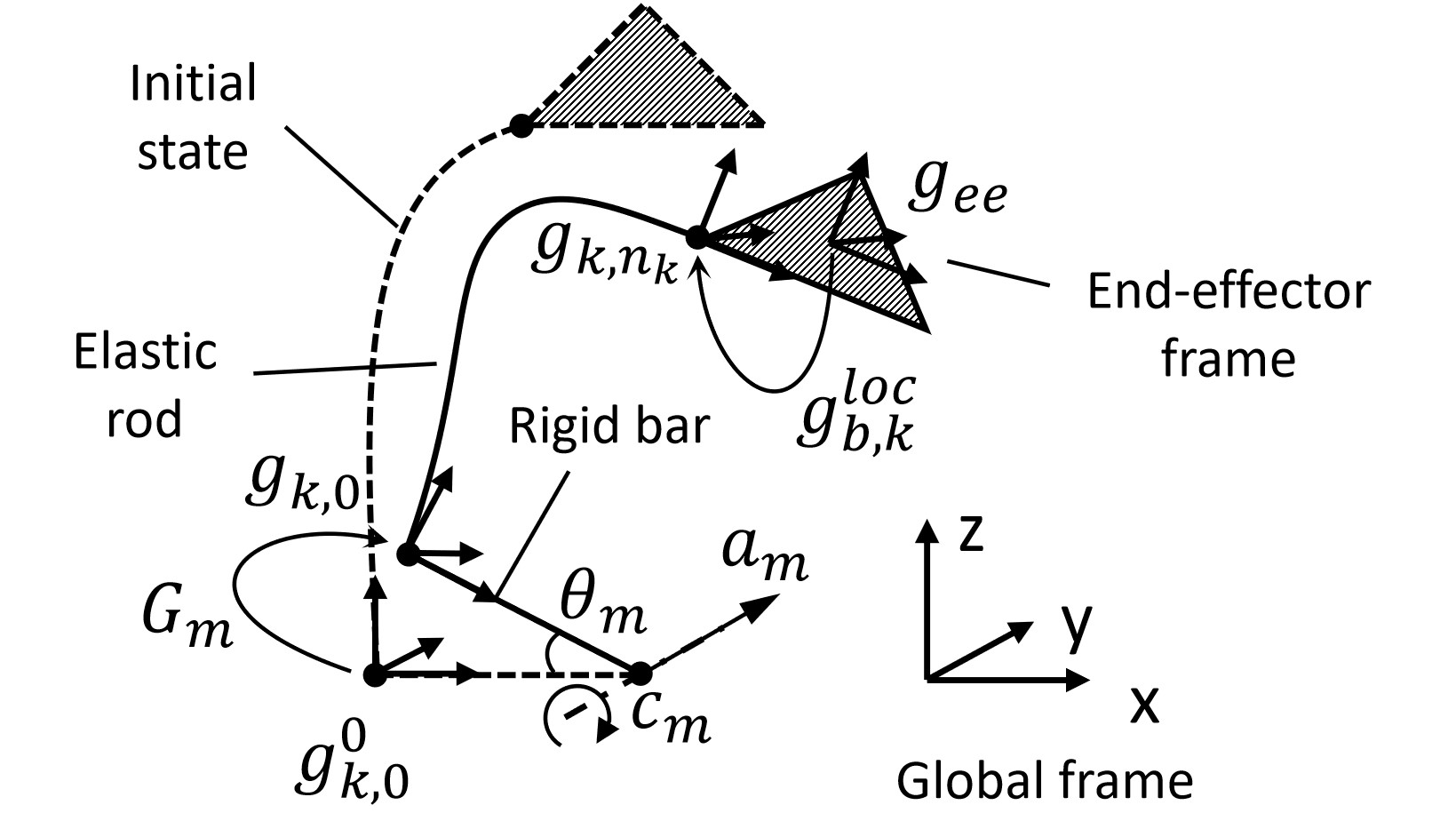}
	\caption{Boundary structure.}
	\label{fig:structure}
\end{figure}

As illustrated in Fig.~\ref{fig:structure}, only the two boundary nodes are prescribed by rigid-body kinematics. At the platform,
\begin{equation}
g_{k,n_k}=g_{\mathrm{ee}}\,g^{\mathrm{loc}}_{b,k},
\qquad
\delta\zeta_{k,n_k}=\operatorname{Ad}_{(g^{\mathrm{loc}}_{b,k})^{-1}}\,\delta\zeta_{\mathrm{ee}},
\label{eq:platform_embed_var_compact}
\end{equation}
where \(g_{\mathrm{ee}}\) is the end-effector pose and \(g^{\mathrm{loc}}_{b,k}\) is a constant attachment transform.

At the base, rod \(k\) is driven by motor \(m\), modeled as a revolute joint about a fixed world axis \((a_m,c_m)\) (unit direction \(a_m\) passing through point \(c_m\)). Denoting by \(g^0_{k,0}\) the installation pose at
\(\theta_m=0\), we define the embedding by the world left action
\begin{equation}
g_{k,0}=\Phi_k(\theta_m):=G_m(\theta_m)\,g^0_{k,0},
\quad
\delta\zeta_{k,0}=S_k(\theta_m)\,\delta\theta_m,
\label{eq:base_embed_var_compact}
\end{equation}
with the right-trivialized Jacobian
\begin{equation}
S_k(\theta_m)=\operatorname{Ad}_{\Phi_k(\theta_m)^{-1}}\xi_m^{s},
\qquad
\xi_m^{s}=
\begin{bmatrix}
a_m\\
-\,a_m\times c_m
\end{bmatrix}.
\label{eq:S_compact}
\end{equation}
here \(G_m(\theta_m)\in SE(3)\) is the homogeneous transform of a rotation by \(\theta_m\) about the world axis
\((a_m,c_m)\), i.e., \(x\mapsto c_m+R(\theta_m)(x-c_m)\).
%
\subsection{Generalized Coordinates and Differential Mapping} \label{subsec:gen_dof_mapping}
The configuration consists of the motor angles, the end-effector pose, and all rod interior variables.
We stack them as
\begin{equation}
q := \big[\ {\theta}\ ;\ g_{\mathrm{ee}}\ ;\ g_{\mathrm{int}}\ ;\ {\beta}\ \big]\ \in\ \mathcal{M},
\label{eq:q_def}
\end{equation}
with the product manifold
\begin{equation}
\mathcal{M}=\mathbb{R}^{3}\times SE(3)^{N_g}\times\mathbb{R}^{6N_\beta}, 
\label{eq:manifold_M}
\end{equation}
$$N_g = 1 + \sum_{k=1}^{6}(n_k-1),\qquad N_\beta=\sum_{k=1}^{6}n_k$$
where \(N_g\) is the total number of independent nodal poses (the end-effector pose plus all rod interior nodes), and \(N_\beta\) is the total number of LSE elements. Here \({\theta}\in\mathbb{R}^3\) collects the motor angles, \(g_{\mathrm{int}}\in SE(3)^{N_g-1}\) stacks all
rod interior nodal poses \(\{g_{k,j}\}_{j=1}^{n_k-1}\) over \(k=1,\dots,6\), and
\({\beta}\in\mathbb{R}^{6N_\beta}\) stacks all strain-slope variables \(\{\beta_{k,e}\}_{e=1}^{n_k}\) over
\(k=1,\dots,6\). The boundary nodes \(\{g_{k,0},g_{k,n_k}\}\) are eliminated by the embeddings in
Section~\ref{subsec:boundary_embed}.

For subsequent linearization, we use right-trivialized increments. The corresponding generalized differential vector is
\begin{equation}
\delta q := \big[\ \delta{\theta}\ ;\ \delta\zeta_{\mathrm{ee}}\ ;\ \delta\zeta_{\mathrm{int}}\ ;\ \delta{\beta}\ \big],
\label{eq:delta_q_def}
\end{equation}
where \(\delta\zeta_{\mathrm{int}}\in\mathbb{R}^{6(N_g-1)}\) and \(\delta{\beta}\in\mathbb{R}^{6N_\beta}\)
are the stacked increments associated with \(g_{\mathrm{int}}\) and \({\beta}\), respectively.

For element \((k,e)\), we introduce a sparse mapping matrix \(P_{k,e}(\theta)\) such that
\begin{equation}
\delta x_{k,e} = P_{k,e}(\theta)\,\delta q.
\label{eq:Pkedef}
\end{equation}

Its block rows simply select the corresponding entries from \(\delta q\), except at boundary nodes:
for an endpoint node \(j\in\{e-1,e\}\),
\begin{equation}\label{eq:P_blocks_cases}
\delta\zeta_{k,j} =
\begin{cases}
\delta\zeta_{k,j}, & j\in\{1,\dots,n_k-1\},\\
S_k(\theta_{m(k)})\,\delta\theta_{m(k)}, & j=0,\\
\operatorname{Ad}_{(g^{\mathrm{loc}}_{b,k})^{-1}}\,\delta\zeta_{\mathrm{ee}}, & j=n_k,
\end{cases}
\end{equation}
and \(\delta\beta_{k,e}\) is always selected by an identity block.

Combining~\eqref{eq:bx} with~\eqref{eq:Pkedef} gives
\begin{equation}
\delta\bar{\xi}_{k,e}
=
B_{k,e}\,\delta x_{k,e}
=
B_{k,e}\,P_{k,e}(\theta)\,\delta q,
\label{eq:delta_barxi_generalized}
\end{equation}
so that rigid--continuum attachments enter only through the mapping~\eqref{eq:P_blocks_cases}, without additional
connection constraints.

In summary, this section establishes the differential kinematic backbone of our formulation. We derived an explicit element-level linearization of the mean strain from~\eqref{eq:explicit_meanstrain} and incorporated rigid--continuum attachments through a sparse mapping from generalized increments to local element perturbations. As a result, the variation of each element can be written directly in terms of \(\delta q\) without introducing additional connection constraints. In the next section, we leverage these Jacobians to derive the total potential energy, assemble the global equilibrium residual, and construct Newton tangent stiffness on the product manifold \(\mathcal{M}\).
%
\section{STATIC EQUILIBRIUM as MANIFOLD OPTIMIZATION}
\label{sec:statics_opt}
This section formulates forward statics as an optimization problem on the product manifold \(\mathcal{M}\). We first derive a closed-form expression for the LSE element energy and its first variation, and then assemble the global equilibrium residual by summing element contributions and external virtual-work terms. To solve the resulting nonlinear system efficiently, we construct explicit Newton tangent and apply a Riemannian Newton iteration with exponential updates on the \(SE(3)\) components.
\subsection{Element Energy} \label{subsec:total_energy}
To simplify the notation, we omit the rod indices \((k,e)\) in the following element-level expressions
(\(U_{\mathrm{int}}, \delta U_{\mathrm{int}}, J_1,J_2,J_3,\Delta,\beta,P\)). The indices are reinstated when the global residual is assembled.
Consider one LSE element of arc length \(h\) with end poses \(g_a=g(0)\), \(g_b=g(h)\) and slope parameter \(\beta\).
Under the linear strain assumption \(\xi(s)=\bar\xi+(s-\tfrac{h}{2})\beta\) and the explicit mean-strain mapping
\eqref{eq:explicit_meanstrain}, the internal elastic energy of the element admits the closed form
\begin{equation}
U_{\mathrm{int}}(q)
=\frac12\,h(\bar\xi-\xi_0)^\top K(\bar\xi-\xi_0)
+\frac{1}{24}\,h^3\,\beta^\top K\beta,
\label{eq:element_energy}
\end{equation}
where \(K\in\mathbb{R}^{6\times 6}\) is the sectional stiffness matrix and \(\xi_0\in\mathbb{R}^6\) is the natural strain.

Let \(\Delta \doteq \bar\xi-\xi_0\). Taking the first variation of \eqref{eq:element_energy} yields
\begin{equation}
\delta U_{\mathrm{int}}
=
h\,\Delta^\top K\,\delta\bar\xi
+\frac{1}{12}h^3\,\beta^\top K\,\delta\beta
\label{eq:delta_Uint}
\end{equation}
Taking \eqref{eq:bx}-\eqref{eq:Be_def} and \eqref{eq:Pkedef} into \eqref{eq:delta_Uint}, we can get
\begin{equation}
    \delta U_{\mathrm{int}} =  \delta x^\top\,\gamma = \delta q^\top\, r 
\end{equation}
where $\gamma$ denotes the elemental residual w.r.t. $\delta x$ and $r$ denotes the elemental residual w.r.t. $\delta q$.
\begin{equation}
    r(q) = P^\top \gamma , \quad \gamma(q) = \begin{bmatrix}
        hJ_1^\top K\Delta  \\ hJ_2^\top K \Delta  \\ hJ_3^\top K \Delta+\frac{1}{12}h^3\, K\beta
    \end{bmatrix}
\end{equation}
%
\subsection{Nodal External Force}
We suppose that the force is applied to the node.
Let \(F_i\in\mathbb{R}^6\) be the external wrenches expressed in body frame applied at the node $i$. For a perturbation $\delta\zeta_i\in\mathbb{R}^6\simeq\mathfrak{se}(3)$,  we define the external load potential through its variation as
\begin{equation}
\delta U_{\mathrm{ext}}^i= -\delta\zeta_i^\top F_i = -\delta q^\top D_i^\top F_i, \quad i\in\{1,\dots,N_g\},
\label{eq:element_ext_vw}
\end{equation}
where $D_i$ is the sparse matrix that extracts $\delta\zeta_i$ from $\delta q$, i.e., $\delta\zeta_i = D_i \, \delta q$.
%
\subsection{System Statics}
The total potential energy of the system is assembled as the sum of element energies:
\begin{equation}
U(q)
=
\sum_{k=1}^{6}\sum_{e=1}^{n_k}
U^{k,e}_{\mathrm{int}}
+
\sum_{i=1}^{N_g}U_{\mathrm{ext}}^i,
\label{eq:total_energy_global}
\end{equation}

The static equilibrium corresponds to a stationary point of \eqref{eq:total_energy_global}. We therefore define the forward statics problem as
\begin{equation}
q^\star
=
\operatorname*{arg\,min}_{q\in\mathcal{M}}
U(q),
\label{eq:static_optimization}
\end{equation}
which is solved in practice by enforcing the vanishing first variation (equilibrium residual) of \(U\). Thus, the residual of the system is
\begin{equation}\label{eq:res}
    r(q) = \sum_{k=1}^{6}\sum_{e=1}^{n_k}P_{k,e}^\top\,\gamma_{k,e}-
\sum_{i=1}^{N_g} D_i^\top F_i=0.
\end{equation}
%
\subsection{Tangent Stiffness}
\label{subsec:tangent_stiffness}
To solve the nonlinear equilibrium equations~\eqref{eq:res} efficiently, we linearize the residual at the current iterate while freezing the kinematic Jacobians \(J_{1,2,3}\) and the mapping \(P\). This amounts to neglecting second-order terms in \(\delta J_i\) and \(\delta P\), thereby yielding explicit tangent operators.
Using~\eqref{eq:linearization}--\eqref{eq:Be_def} and retaining only first-order terms through \(\delta\bar\xi\) and \(\delta\beta\), the linearization takes the form
\begin{equation}
\delta \gamma = H\,\delta x,
\label{eq:local_tangent}
\end{equation}
with the symmetric block matrix
\begin{equation}
H =
\begin{bmatrix}
hJ_1^\top KJ_1 & hJ_1^\top KJ_2 & hJ_1^\top KJ_3\\
hJ_2^\top KJ_1 & hJ_2^\top KJ_2 & hJ_2^\top KJ_3\\
hJ_3^\top KJ_1 & hJ_3^\top KJ_2 & hJ_3^\top KJ_3 + \frac{1}{12}h^3K
\end{bmatrix}.
\label{eq:Hx_blocks}
\end{equation}

Combining \eqref{eq:Pkedef} with \eqref{eq:local_tangent} yields the global tangent stiffness:
\begin{equation}\label{eq:K}
    K_{\mathrm{t}}(q)=\sum_{k=1}^6\sum_{e=1}^{n_k} P_{k,e}^\top \,H_{k,e} \,P_{k,e}.
\end{equation}
%
\subsection{Riemannian Newton Solver} \label{subsec:global_solver}
To solve \eqref{eq:res}, we employ a Riemannian Newton iteration \cite{sato2021riemannian} on the product manifold \(\mathcal{M}\).
At each iteration, we compute a tangent increment \(\delta q\in T_q\mathcal{M}\) by solving the linearized system
\begin{equation}
K_{\mathrm{t}}(q)\,\delta q
=
-\,r(q).
\label{eq:gn_step_statics}
\end{equation}

The configuration is then updated using exponential retractions on the \(SE(3)\) components and additive updates on the Euclidean components:
\begin{equation}
g \leftarrow g\,\exp(\widehat{\delta\zeta}),\qquad
{\beta} \leftarrow {\beta} + \delta{\beta},\qquad
{\theta} \leftarrow {\theta} + \delta{\theta},
\label{eq:update_rule_statics}
\end{equation}
applied to the end-effector pose and all rod interior nodes (\(g\in SE(3)\)), as well as to all strain-slope and motor-angle variables. The iteration terminates when \(\|r(q)\|\) falls below a prescribed tolerance.

\begin{remark}
When the motor angles \(\theta\) are prescribed, they are treated as parameters rather than unknowns. The equilibrium residual and tangent stiffness are reduced by removing the rows and columns associated with \(\theta\), leaving a system in the remaining variables.
\end{remark}

In summary, we cast the forward statics of the continuum parallel robot as a manifold equilibrium problem and derived
an explicit, assembly-ready formulation: closed-form element energies, global residuals obtained by virtual work, and Newton tangent projected through the rigid--continuum embeddings. This yields an efficient Riemannian Newton solver that computes the end-effector pose and rod deformations under given motor inputs and
external loads. In the next section, we validate the proposed model experimentally by comparing predicted
end-effector trajectories and configurations against measurements on the physical prototype, both without and with external loading.
%
\section{EXPERIMENTAL VALIDATION}
\label{sec:experiments}
This section presents the experimental validation of the proposed static model. The goal is to assess whether the model can reliably predict the equilibrium behavior of the continuum parallel robot under representative operating conditions. To this end, we perform hardware experiments and reproduce the same conditions in simulation, and then compare the resulting end-effector motion and robot configurations.
The validation is organized into two groups of experiments: a baseline case without external load and a loaded case with an external force applied to the end-effector. 
\begin{figure}[h]
	\centering
    \includegraphics[width=0.49\textwidth]{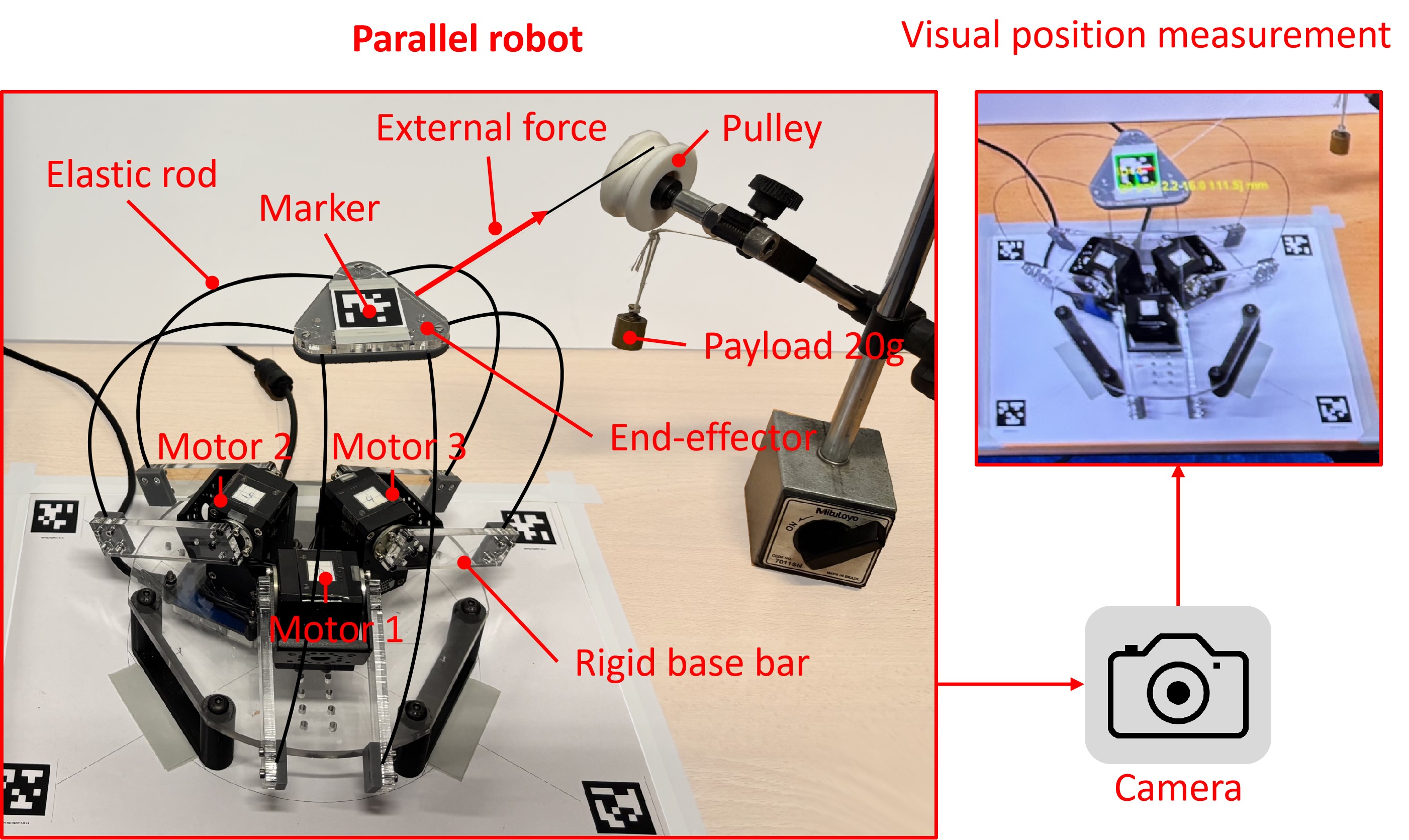}
	\caption{Experimental setup for validating the continuum parallel robot static model. }
	\label{fig:setup}
\end{figure}
%
\subsection{Experimental Platform and Validation Objective}
\label{subsec:exp_platform}
The experimental platform is a continuum parallel robot composed of three servomotors mounted on a rigid base and six elastic rods (two rods per motor) connected to a common end-effector, as shown in Fig.~\ref{fig:setup}. A visual position measurement setup is used to track the end-effector motion during experiments, in which a marker attached to the end-effector is imaged by a camera. For the loaded tests, an external force is applied via a rope-pulley-mass arrangement, which provides a simple and repeatable means to generate a tensile load on the end-effector.

The objective of this section is to validate the proposed static model by comparing experiments and simulations under representative conditions. We assess whether the model reproduces the end-effector equilibrium motion and overall robot configuration in free motion and under external loading. Identical actuation inputs are applied in hardware and simulation, and end-effector positions and configurations are compared at matched phase samples.
%
\subsection{Actuation Protocol and Phase-Sampled Trajectory}
\label{subsec:actuation_protocol}

To generate a repeatable closed-end-effector motion for validation, the three base motors are driven by sinusoidal inputs with a \(2\pi/3\) rad phase shift. The motor angles are prescribed as
\begin{equation}
\boldsymbol{\theta}(t)=
-\frac{\pi}{12}
\begin{bmatrix}
\sin(\omega t)+1\\
\sin(\omega t+\frac{2\pi}{3})+1\\
\sin(\omega t+\frac{4\pi}{3})+1
\end{bmatrix},
\label{eq:actuation_protocol}
\end{equation}
where \(\boldsymbol{\theta}(t)=[\theta_1(t),\theta_2(t),\theta_3(t)]^\top\) and \(\omega\) is the actuation angular frequency (set to \(0.5\) rad/s). This three-phase excitation produces a periodic end-effector motion with an approximately closed circular trajectory.

In both experiments and simulations, the robot starts from the zero-angle configuration, which defines the first sample. The motors then ramp at constant speed to the sinusoidal actuation in~\eqref{eq:actuation_protocol}; the end of this transition is the second sample. The subsequent motion is periodic. Over one period, we select ten uniformly spaced phase samples, with the first phase coinciding with the transition endpoint, yielding 11 samples in total. At each sample, we record the end-effector position and capture a photo for qualitative comparison with the simulated configuration. The same phase-aligned protocol is used in both unloaded and loaded tests.
We first measured the end-effector position of the continuum parallel robot under different loads and motor angles. By comparing experimental measurements with simulation results, the stiffness parameters of the elastic rods were identified. 
In the simulation, each rod was discretized into four LSE elements with a diameter of \(1\,\mathrm{mm}\) and a Young's modulus of \(1.13\times10^{10}\,\mathrm{Pa}\).%
\subsection{Experiment I: Validation Without External Load} \label{subsec:exp_free}
In Experiment~I, no external load is applied. The robot is driven only by the three-phase motor actuation in \eqref{eq:actuation_protocol}, providing a baseline validation of the proposed static model under actuation-induced deformation.
\begin{figure}[h]
	\centering
    \includegraphics[width=0.45\textwidth]{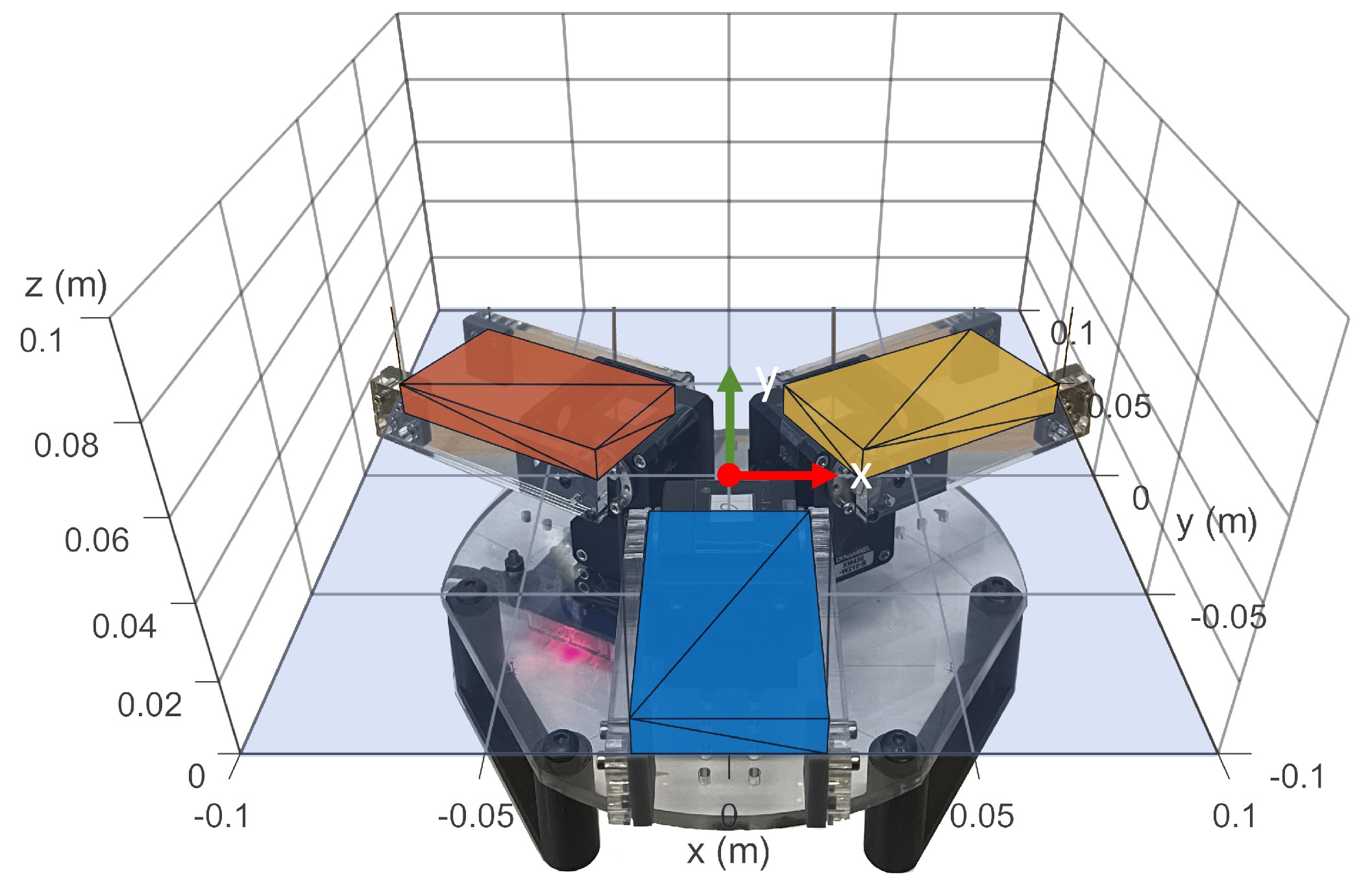}
	\caption{The global frame used for experiments and simulations.}
	\label{fig:frame}
\end{figure}

A global frame is defined for both measurement and simulation. Its origin lies on the robot central axis, within the plane spanned by the three motor rotation axes (Fig.~\ref{fig:frame}). End-effector positions are expressed in this common frame.
One actuation period is uniformly sampled at ten phases. The same motor inputs and phase samples are used in simulation to compute the corresponding equilibrium states, enabling comparison of end-effector positions and shape evolution along the trajectory. 
%
%
\begin{figure}[h]
  \centering
  \begin{subfigure}[h]{1\linewidth}
    \centering
    \includegraphics[width=\linewidth]{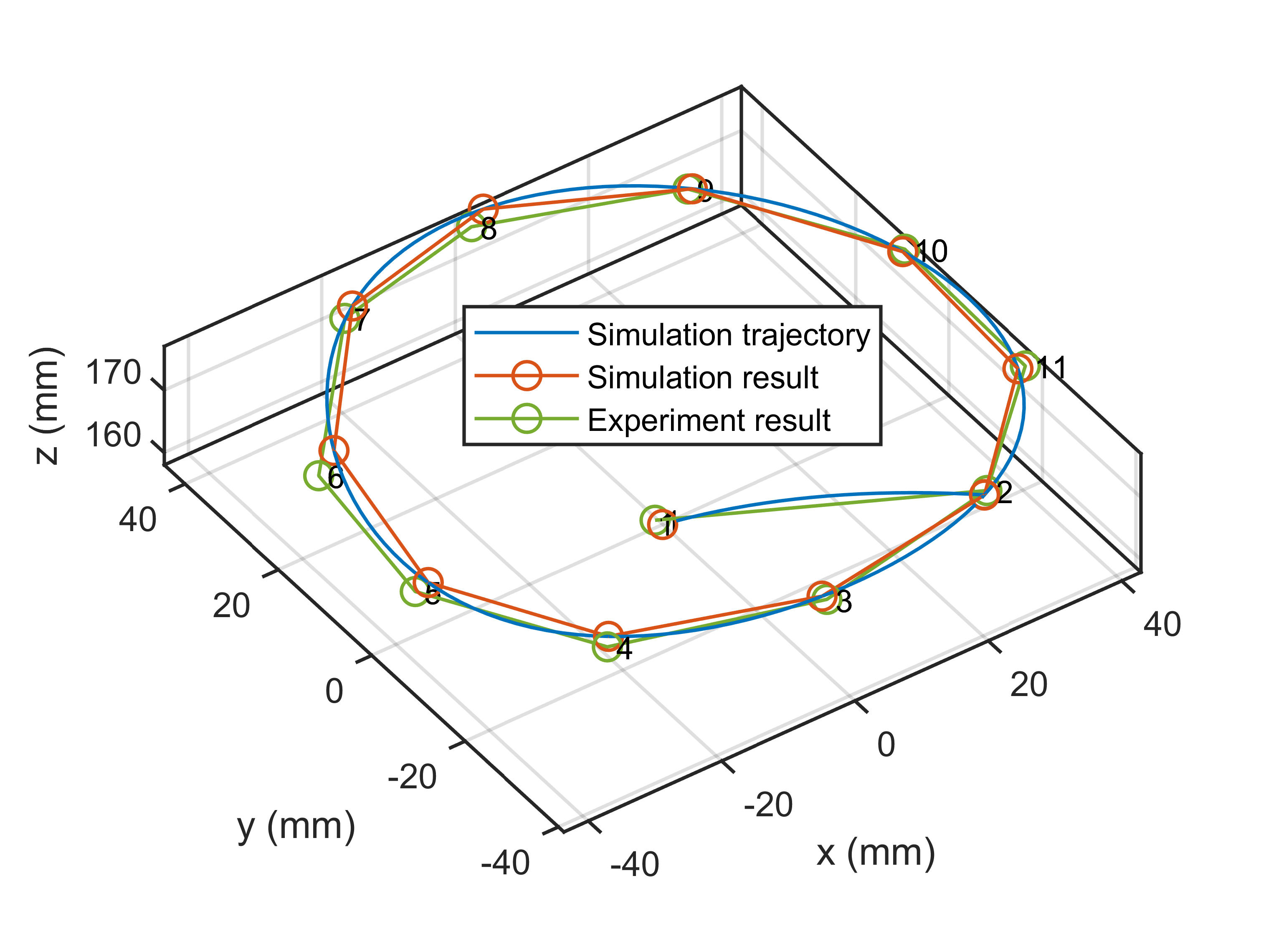}
    \caption{Without external force}
    \label{fig:xxx_a}
  \end{subfigure}
  
  \vspace{0.3em}
  \begin{subfigure}[h]{1\linewidth}
    \centering
    \includegraphics[width=\linewidth]{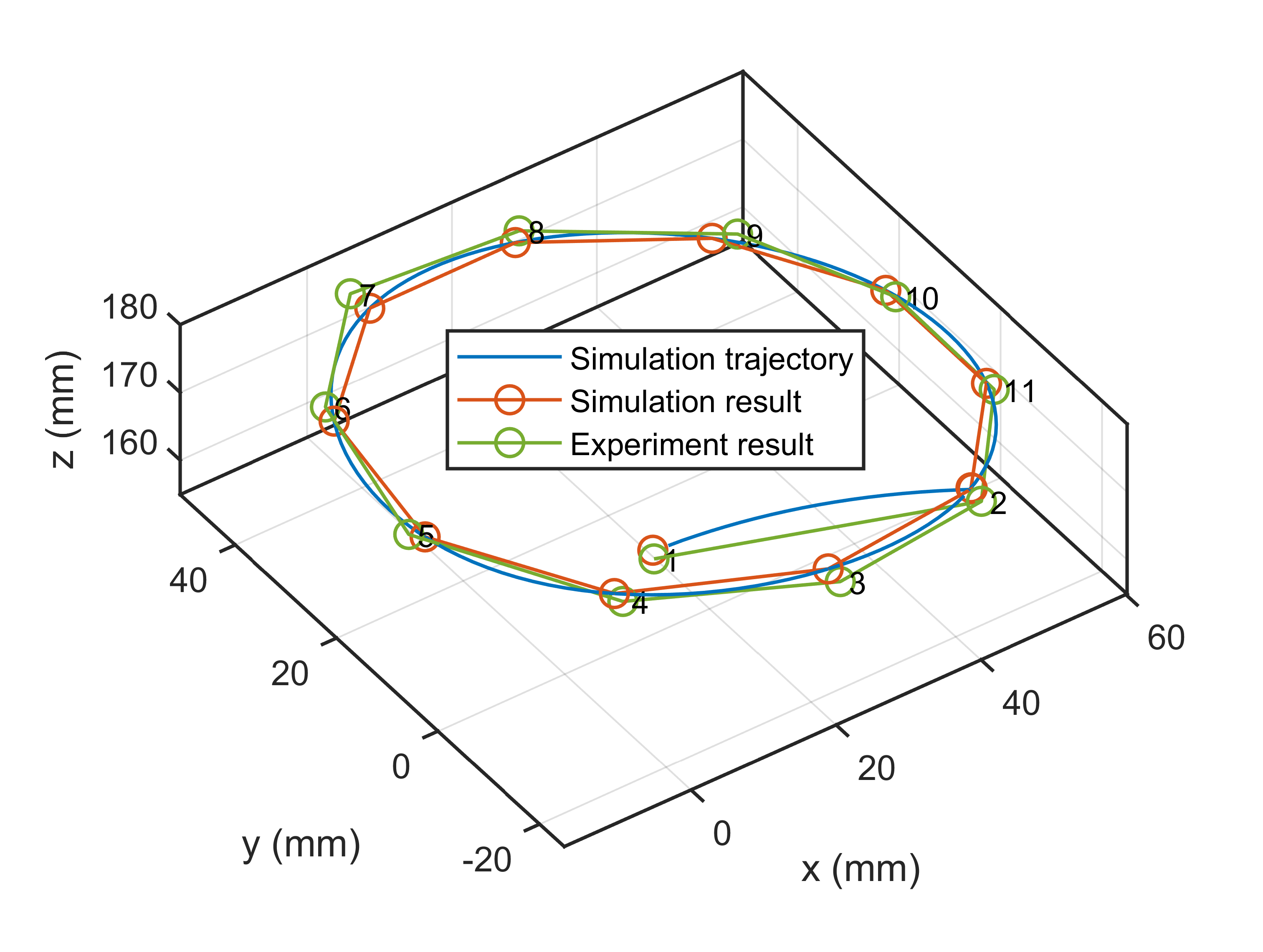}
    \caption{With external force}
    \label{fig:xxx_b}
  \end{subfigure}
  \caption{Comparison of end-effector trajectories and sampled positions between simulation and experiment. }
  \label{fig:position}
\end{figure}
\subsection{Experiment II: Validation Under External Tension Load} \label{subsec:exp_loaded}
Experiment~II uses the same actuation~\eqref{eq:actuation_protocol} and ten-phase sampling, but applies an external tensile load via a rope--pulley--mass setup. The load magnitude is approximately constant (payload weight), while its direction varies with the end-effector position and always points toward the pulley. In simulation, we impose the corresponding force at the same pulley location and payload magnitude, and compare the results with experiments to assess load-induced changes in equilibrium motion and configuration.
\subsection{Simulation--Experiment Comparison Metrics}
\label{subsec:metrics}

For both the unloaded and loaded experiments, the simulation results are compared with the corresponding measurements at the same 11 sampling points. Let \(\mathbf{p}_k^{\mathrm{exp}}\) and \(\mathbf{p}_k^{\mathrm{sim}}\) denote the measured and simulated end-effector positions at the \(k\)-th sampling point, respectively, expressed in the global coordinate frame. The pointwise position error is defined as
\begin{equation}
e_k = \left\| \mathbf{p}_k^{\mathrm{exp}} - \mathbf{p}_k^{\mathrm{sim}} \right\|_2,
\quad k=1,\dots,11 .
\label{eq:pos_error}
\end{equation}

Based on \(\{e_k\}_{k=1}^{11}\), we report the mean error and maximum error for each experiment. In addition to the position-based metrics, a qualitative comparison is performed by visually inspecting robot photos and simulated configurations at matched sampling points to assess whether the model captures the robot's overall shape evolution.
\begin{figure*}[t]
  \centering
  \begin{subfigure}[t]{1\linewidth}
    \centering
    \includegraphics[width=\linewidth]{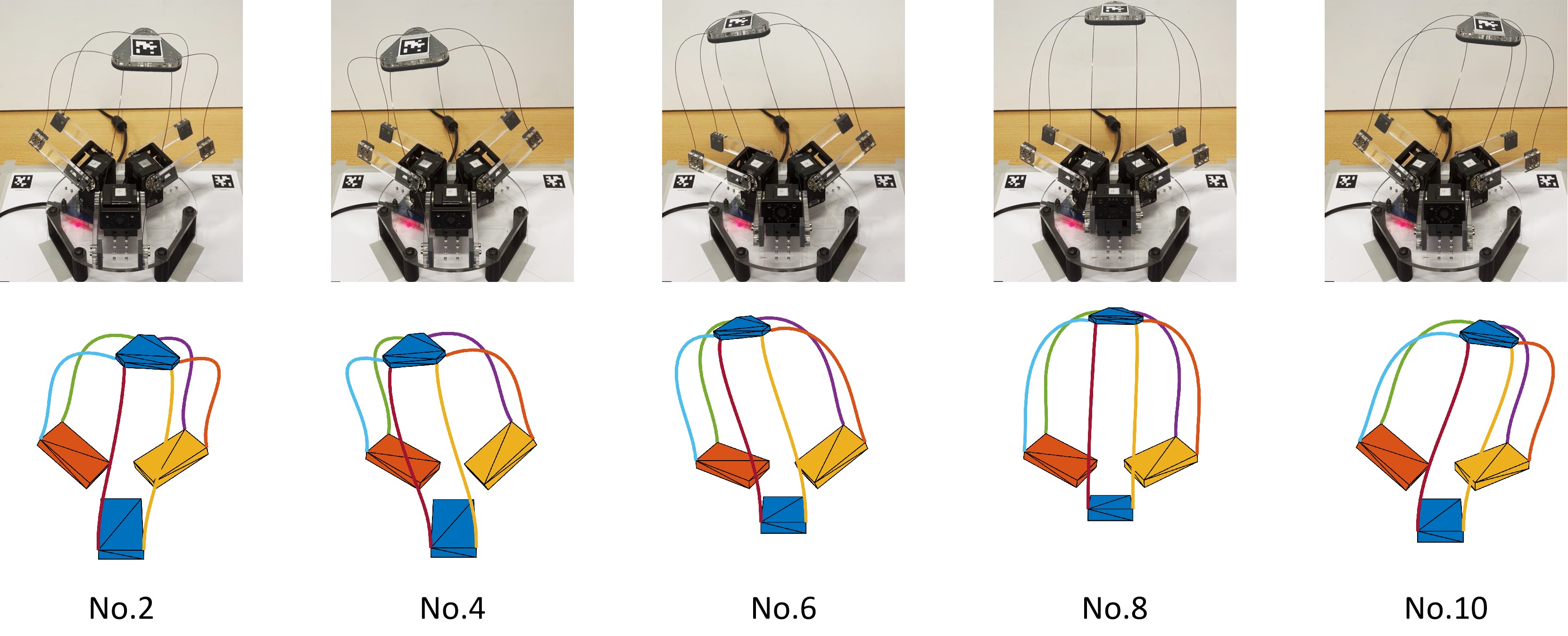}
    \caption{Without external force}
    \label{fig:cxxx_a}
  \end{subfigure}
  \vspace{0.5em}
  
  \begin{subfigure}[t]{1\linewidth}
    \centering
    \includegraphics[width=\linewidth]{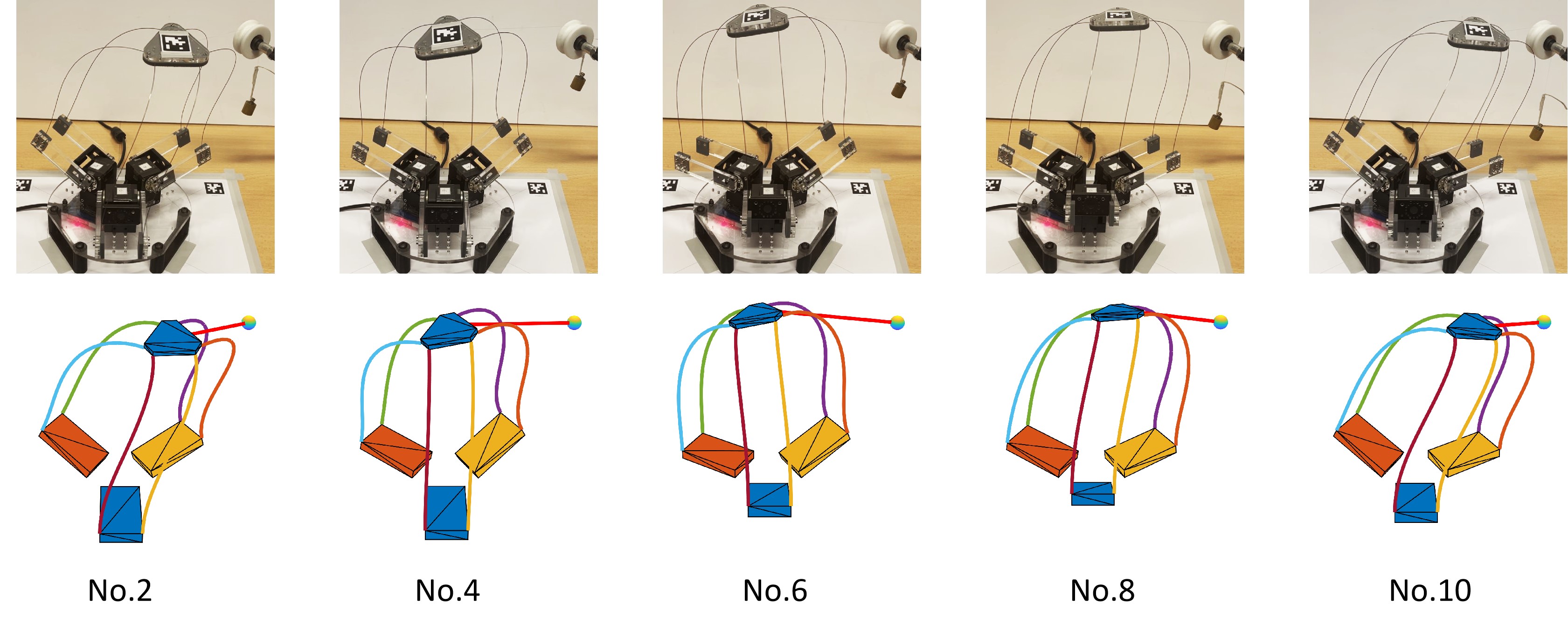}
    \caption{With external force}
    \label{fig:cxxx_b}
  \end{subfigure}
  \caption{Qualitative comparison of robot configurations between experiment (top row) and simulation (bottom row) at representative sampling points. In simulation, each rod is discretized with \(4\) LSE elements.}
  \label{fig:configuration}
\end{figure*}
%
\subsection{Results and Discussion}
\label{subsec:results_discussion}

We compare simulations and experiments in both unloaded and loaded conditions. For each case, we report the end-effector trajectory using the 11 sampled positions (Fig.~\ref{fig:position}) and visually compare representative configurations at matched indices (Fig.~\ref{fig:configuration}). 
\begin{table}[h]
\centering
\caption{Position error statistics of simulation--experiment comparison at 11 sampling points.}
\label{tab:error_metrics}
{%
\renewcommand{\arraystretch}{1.2}
\begin{tabular}{lcc}
\hline
Metric & Without external force & With external force \\
\hline
Mean error (mm) & 2.1 & 3.5\\
Max error (mm)  & 3.8 & 4.2\\
\hline
\end{tabular}
}%
\end{table}

\subsubsection{Free-motion case} \label{subsubsec:results_free}

In the unloaded case, the simulated configurations match the photos in overall deformation patterns, including platform inclination and rod bending trends (Fig.~\ref{fig:cxxx_a}). The end-effector trajectory is approximately closed and circular, and the phase ordering of the sampled points is consistent between experiment and simulation (Fig.~\ref{fig:xxx_a}). The sampled positions agree closely with the simulated trajectory; quantitative errors are reported in Table~\ref{tab:error_metrics}.
\subsubsection{Loaded case with pulley-induced tension}
\label{subsubsec:results_loaded}
With the pulley-induced tension, the end-effector trajectory exhibits a clear load-induced shift and shape change relative to the unloaded case (Fig.~\ref{fig:xxx_b}). The simulation reproduces this trend and remains consistent with the measured samples, indicating that the model captures the coupled effect of actuation and external loading. Representative configuration comparisons also show similar load-biased deformations at matched indices (Fig.~\ref{fig:cxxx_b}). The remaining discrepancies are primarily due to unmodeled effects, including pulley friction, rope alignment error, and assembly tolerances (Table~\ref{tab:error_metrics}).

Overall, the experiments validate the proposed static model in both unloaded and externally loaded conditions, supporting its use for predicting equilibrium motion and configuration of the continuum parallel robot.
%
\section{CONCLUSION}\label{sec.conc}
This paper presented a compact, Lie-group-consistent forward-statics formulation for continuum parallel robots that avoids explicit rigid--continuum connection constraints. Each rod is discretized by nodal poses on \(SE(3)\), while an element-wise strain field is reconstructed locally via a linear strain element. Using a fourth-order Magnus approximation, we derived an explicit and geometrically consistent pose--strain mapping, eliminating inner strain-recovery iterations. Rigid attachments at the servomotor-driven base and the end-effector platform are enforced by construction through kinematic embeddings and a sparse differential projection from generalized increments to element perturbations. Based on total potential energy and virtual work, we obtained an assembly-ready equilibrium residual and an explicit Newton tangent, enabling an efficient Riemannian Newton iteration on the product manifold of pose and strain variables.
Experiments on a three-servomotor, six-rod prototype validated the proposed model in both unloaded motions and externally loaded cases with pulley-induced tension, showing good agreement between simulated and measured end-effector trajectories and representative configurations. These results suggest that the proposed formulation provides a practical, control-friendly basis for equilibrium prediction in closed-chain rod-based mechanisms.
Future work will extend the framework to dynamics and to model-based real-time state estimation and control of CPRs.

\appendix[]
\subsection{Adjoint Representation of the Lie Algebra and Lie Group}\label{notations}
	The adjoint representation of the Lie algebra $\mathfrak{se}(3)$ is 
	$$
	{\rm{ad}}_{{\xi}}= \left(\begin{matrix}
		\tilde{{\kappa}}&{0}\\\tilde{{\epsilon}}&\tilde{{\kappa}}
	\end{matrix}\right)\in \mathbb{R}^{6\times6}, \quad \xi=[\kappa^\top \ \epsilon^\top]^\top\in \mathbb R^6.
		$$
        
        The adjoint representation of the Lie group $SE(3)$ is 	
		$${\rm{Ad}}_{{g}}= \left(\begin{matrix}
			{R}&{0}\\\tilde{{p}}{R}&{R}
		\end{matrix}\right)\in \mathbb{R}^{6\times6}.$$

\subsection{Tangent Operator}
For $x\in \mathfrak{se}(3)$, the tangent operator of the exponential map and its inverse are defined as follows~\cite{hairer2006geometric}:
\begin{equation}
	\mathrm{dexp}_x = \sum_{j=0}^{\infty} \frac{1}{(j+1)!} \, \mathrm{ad}^j_x, \quad 
	\mathrm{dexp}_x^{-1} = \sum_{j=0}^{\infty} \frac{B_j}{j!} \, \mathrm{ad}^j_x
\end{equation}
where \( B_j \) are the Bernoulli numbers.

\bibliographystyle{IEEEtran}
\bibliography{biblio.bib}

\end{document}